\documentclass[a4paper,11pt]{article}

\usepackage{breakurl}
\usepackage[breaklinks=true]{hyperref}
\usepackage{breakcites}
\usepackage{multirow}
\usepackage{amssymb,amsmath,array,bm}
\usepackage{booktabs}

\usepackage{tikz}
\usetikzlibrary{arrows}

\usepackage{pgfplots}
\pgfplotsset{compat=1.15}
\usetikzlibrary{arrows}

\usepackage[a4paper, left=3cm, right=2.5cm, top=3cm, bottom=3cm]{geometry}

\usepackage{authblk}
\usepackage{natbib}
\usepackage{doi}

\usepackage[utf8]{inputenc}
\usepackage[T1]{fontenc}
\usepackage[english]{babel}
\usepackage{csquotes}

\title{Prototype Adaptation for Zero-Shot sEMG Movement Classification}

\author[1]{Rui Liu}
\author[1]{Benjamin Paaßen}
\affil[1]{Faculty of Technology, Bielefeld University}
\affil[1]{Faculty of Technology, Bielefeld University}

\date{27.07.2026}

\begin{document}

\maketitle

\pagestyle{myheadings}
\markright{preprint as provided by the authors}

\begin{abstract}
Surface electromyography (sEMG) enables the control of prostheses, allowing upper-limb amputees to re-gain some hand function. Most current research focuses on recognizing basic movements for prosthesis control. However, in most daily activities, such as opening a door, combined movements are essential. However, collecting training data for all possible combined movements is time-consuming and requires re-training of the model for any new combination. We propose two novel recognition approaches, Compositional Prototype Interpolation (CPI) and Synthetic Adaptation for Prototypes (SAP), that enable zero-shot learning of combined, novel and unseen movements in Prototype Networks after training only with basic movements. Our methods rest on a linear interpolation assumption in the embedding space, which we study by inspecting the geometry of combined motions in signal and embedding space. In experiments on the NearLab and NinaPro DB3 data sets as well as our newly recorded BasCom dataset, our proposed SAP outperforms prior zero-shot learning methods with accuracy improvements on combined movements of more than 20\%. This advantage is maintained in online inference experiments in a user study. 

\end{abstract}

\section{Introduction}
\label{sec:introduction}
In recent decades, surface electromyography (sEMG) has been widely applied for rehabilitation and prosthesis control \cite{sturma2018rehabilitation}.
Recently, deep learning methods have improved movement classification accuracy compared to pre-existing shallow models \cite{li2021gesture}.
However, deep models require large amounts of sEMG training data for every movement that shall be recognized, and the model has to be re-trained whenever a new movement is added.
In clinical application, this need for repeated data collection and re-training is known as the "calibration burden" \cite{wang2024robust}.

A promising approach to reduce this burden is zero-shot learning (ZSL), meaning the classification of new movements based only on training data for a base set of movements. The foundation for ZSL in movement classification is muscle synergies theory, which suggests that combined movements can be expressed as linear combinations of basic synergies 
\cite{tresch2006matrix,bizzi2013neural}.
Following this theory, 
we assume that, in a ideal representation space, a combined motion should be a convex combination of its constituent movements. Accordingly, the signal representing the unseen combined movement can be approximated by linear interpolation between the signals of its constituent movements.

Several prior works have attempted to utilize such linear interpolation approaches, typically generating a sample of a combined movement by a 50\%/50\% mixture of samples of sEMG signals for basic movements.
Unfortunately, the linear interpolation assumption almost never holds exactly---e.g., due to the physical constraints of signal transmission, such as crosstalk and noise.
Non-linearities are particularly severe 
for amputees in clinical applications, 
e.g.\ because amputees employ strategies to compensate for a missing hand by utilizing the arm more, leading to different sEMG signals 
\cite{hussein2026upper}. 
Overall, ZSL based on current signal mixup approaches achieves limited performance on combined movements.

To improve ZSL performance, we propose two novel strategies: (1) We represent combined movements via interpolated prototypes in a learned embedding space of a prototype network (Compositional Prototype Interpolation; CPI). (2) We adapt such prototypes in the embedding space using only synthetic data, achieving decision boundaries that remain robust to some violations of the linearity assumption (Synthetic Adaptation for Prototypes; SAP).



In summary, we address the problem of signal non-linearity and calibration burden for prosthetic control with the following contributions: 
(1) We provide a geometrical analysis in sEMG signal space as well as the embedding space of prototype networks, verifying that combined movements are close to a convex combination of basic movements in embedding space. 
(2) We perform interpolation in the embedding space of prototype networks which allows us to create new prototypes for unseen, combined movements after training (CPI). 
(3) We propose a novel synthetic adaptation for prototypes (SAP) method to adjust the location of prototypes for unseen combined movements via synthetic data.
(4) We create a novel dataset (BasCom) containing 11 basic movements and 8 combined movements of $N=11$ participants, which is larger than comparable, prior datasets \cite{soroushmojdehi2022transfer}. 
(5) We evaluate our method on sEMG data from able-bodied participants (NearLab and BasCom data sets), amputees (NinaproDB3 data set), and in an online user study with $N=11$ able-bodied participants.

The results demonstrate the robustness and effectiveness of our methods and verify the potential for clinical application. All data and software is available at \url{https://gitlab.com/semg-zeroshot}.

\section{Related Work}
\paragraph{sEMG-based movement classification}
Using machine learning methods, intended hand movements can be recognized based on biomedical signals, such as near-infrared spectroscopy (NIRS), inertial measurement unit (IMU), and mechanomyographic (MMG) signals. The most common signal source for recognition of hand movement is sEMG \cite{oskoei2007myoelectric}.

Recent research on sEMG-based movement recognition has focused on deep learning methods for hand movement recognition, in particular convolutional neural networks (CNNs). For example, \cite{shahid2022performance} indicates that 1D-CNN requires less computing resources compared to 2D-CNNs, which makes this convolutional layer more suitable for deployment on edge systems like prostheses. \cite{Liu2024} investigates 1D-CNNs for few-shot learning of new movements and \cite{Liu2025} investigates transfer learning schemes against electrode shifts with 1D-CNNs. In this paper, we also utilize a 1D-CNN backbone for movement classification but will expand it for zero-shot classification of new movements.

\paragraph{Zero-shot learning}

Current prostheses control requires complex calibration for new users and movements, such that users must provide a large amount of sEMG signals for training. By contrast, Zero-shot learning (ZSL) would enable the recognition of unseen users or movements without specific calibration.

In recent years, zero-shot learning has achieved remarkable results in computer vision and nature language processing based on natural language-based in-context learning, such as textual descriptions of visual objects. However, such approaches are not applicable for the description and recognition of movements from sEMG data because foundation models for sEMG data that could be conditioned on text are unavailable.

To address the cross-user zero-shot problem, Wang et al. developed ReactEMG \cite{wang2025reactemg}, allowing new users to control prostheses without calibration. Another approach \cite{eddy2024big} by Eddy et al.\ trained models on a large dataset of 612 subjects allowing cross-user myoelectric control without per-user calibration. This work is not applicable in our case because we focus on zero-shot classification of new movements, not generalization to new users.

For the zero-shot problem for new movements, \cite{wang2025new} proposed SeqEMG-GAN to generate synthetic sEMG signals based on sequences of known joint angles in the hand and wrist and train classifiers based on the synthetic data. While this methods has some empiric success, it is limited by the reliance on joint angle data, which are not avialable in our setting.

Muscle synergies theory assumes that the neural motor system constructs a combined movement by linear combination of its basic synergies \cite{bizzi2013neural}. 
Based on this theoretical assumption, several ZSL methods for the recognition of unseen combined movement have been proposed. 
\cite{soroushmojdehi2022transfer} introduced Syn0net for the classification of combined motions by training the model simply on basic movements and classifying high activations for two basic movements as their combination.
Similarly, \cite{yazawa2025recognition} propose a probabilistic scheme to recognize pairwise combinations $k\&l$ of basic movements $k$ and $l$ via soft labeling. The classification is performed by minimizing the Kullback-Leibler divergence between predicted probabilities and synthetic soft labels with pairwise combinations where $y_k = y_l = 0.5$.

\paragraph{Signal mixup for synthetic sEMG data} 
To generate synthetic sEMG signals for combined movements, signal mixup approaches based on a linear interpolation assumption are widely applied.
Input mixup \cite{zhang2017mixup} interpolates raw signals in the input space to generate unseen combined motions. 

A fundamental challenge of such signal mixup approaches lies in the discrepancy between the linear neural motor control pattern and non-linearities that affect the recorded sEMG signals.
In particular, the representation space of recorded sEMG signals is distorted by different physical factors, such as muscle crosstalk and noise during signal transmission. Thus, the recorded raw sEMG from signal space are highly non-linear and does not conform to the above mentioned linear addition of basic movement\cite{oskoei2007myoelectric}.
The challenge is even more severe for amputees.
Due to the physical loss of a hand, clinical studies have noted that the residual proximal arm muscle of amputees often dominate the sEMG signal when attempting combined movements, obscuring concurrent hand activity. Thus, movement classification approaches with high accuracy on able-bodied subjects may not generalize to amputees\cite{atzori2014electromyography}.

As an alternative to signal mixup in the input space, 
manifold mixup has been proposed \cite{verma2019manifold} which performs the interpolation in the embedding space of a deep neural network.
A theoretical basis for interpolating in the embedding space rather than the input space is provided by manifold disentanglement. \cite{brahma2015deep} argue that deep neural networks tend to unfold individual class manifolds, measured by a decrease of manifold curvature for deeper layers. This effect appears both for supervised and unsupervised machine learning, indicating that it is due to network depth itself.


\cite{yazawa2025recognition} compared mixup both in the input and in the embedding space with an MLP classifier and found that mixup in the embedding space performs worse. This is, perhaps, because the embedding space of an MLP is too simplistic and the hidden layers are unstable during training.
The authors recommend improving synthetic data accuracy by projecting signals to a better, non-linear embedding space, enabling better data generation competence. This is one of our goals in this work.



\paragraph{Prototype networks}

Among different deep learning architectures, prototype networks (PN) \cite{snell2017prototypical} are particularly interesting for few-shot settings. The PN method assumes that each class can be represented in embedding space by a single prototype and that samples can be classified via nearest-neighbor classification using these prototypes.
However, PNs traditionally use the class mean to compute the prototype, which may be unduly influenced by outliers and suboptimal for classification. In order to get a better decision boundary, \cite{sharma2023learning} proposes a Learning Prototype Classifier. It treats prototypes as trainable parameters of the model, adjusting the locations of prototypes via backpropagation. It is noted that learning vector quantization approaches have trained prototypes for decades before \cite{Sato1995,Paassen2018Neurocomputing}. 



\section{Method}
This work develops novel zero-shot, incremental learning methods to classify unseen combined movements without forgetting the basic movements in the training data. In more detail, we provide the following algorithmic contributions:

\begin{itemize}
\item \textbf{Geometric Analysis:} We provide a geometrical analysis of the sEMG embedding space, measuring the non-linearity of combined movements in relation to basic movements, either in signal space and embedding space. 
It provides the theoretical basis for the following embedding space interpolation.
\item \textbf{Compositional Prototype Interpolation (CPI):} We propose a novel interpolation method named Compositional Prototype Interpolation (CPI). Different from Mixup methods \cite{zhang2017mixup,verma2019manifold}, we generate synthetic samples and new prototypes in the learned embedding space of a convolutional neural network, enabling zero-shot recognition. This approach eliminates the need for synthetic samples during training as the prototypes can be interpolated after training and directly be used for inference.
\item \textbf{Synthetic Adaptation for Prototypes (SAP):} We develop a novel Synthetic Adaptation for Prototypes (SAP) scheme, which optimizes the locations of prototypes for classification using the synthetic data.
\item \textbf{Baseline Extension (Syn0net+):} To make the baseline approach of Syn0net \cite{soroushmojdehi2022transfer}
comparable to our work, we extend it to classify both basic and combined movements.
\end{itemize}

We begin by describing the geometrical analysis of representation space as well as the CNN backbone of our architecture, and then introduce CPI, SAP, and Syn0net in turn.

\subsection{Geometrical Analysis}
\label{sec: geometric}

The foundational assumption behind our approach is that combined movements can be regarded as convex combinations of their constituent basic movements in the embedding space of a convolutional neural network. More precisely, let $\phi$ denote the convolutional neural network, and let $\bm{x}_k$ and $\bm{x}_l$ be samples of basic movements $k$ and $l$. Then, we assume that a sample of the combined movement $k \& l$ can be generated as
\begin{equation}
\label{eq:linear_combination}
\phi(\bm{x}_{k \& l}) \approx \alpha \cdot \phi(\bm{x}_k) + (1-\alpha) \cdot \phi(\bm{x}_l)
\end{equation}
for an appropriate mixing coefficient $\alpha \in [0, 1]$. In geometric terms, we make two assumptions: (1) Convexity Assumption: The mixing ratio $\alpha$ should lie within the convex interval $[0, 1]$, meaning that the combined movement lies \emph{between} the basic movements in embedding space. 
(2) Linearity Assumption: The linear interpolation assumes that the combined movement lies approximately on a line connecting the basic movements. 

However, if the actual data manifold is highly curved, linearly interpolated signals will deviate from the true distribution. Therefore, the representation space should be sufficiently flattened to ensure that combined movements are close to linear combinations of their constituent basic movements.
\begin{figure}
\includegraphics[width=8cm]{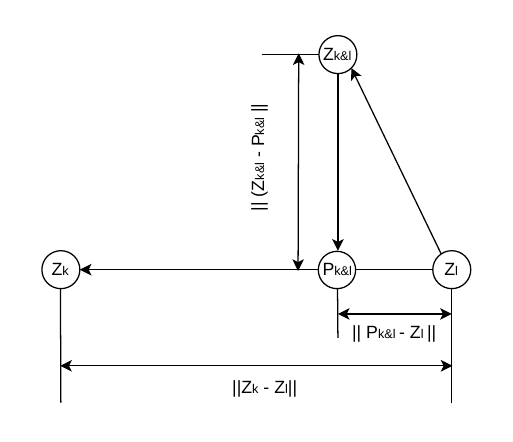}
\centering
\caption{An illustration of the geometry between two basic movements K and L and a combined movement KL.} 
\label{fig:ratio_nonlinear}
\end{figure} 

To quantify how closely data conforms to the convexity and linearity assumption respectively, we propose two metrics, namely the mixing ratio $\alpha$ and the linearity error $\delta$. In more detail, let
$\bm{z}_k$ and $\bm{z}_l$ denote the prototypes of the movements $k$ and $l$, let $\bm{z}_{k \& l}$ denote the prototype of the combined movement $k \& l$, and let $\bm{p}_{k \& l}$ be the projection of $\bm{z}_{k \& l}$ onto $\bm{z}_l - \bm{z}_k$. Then, we define the mixing ratio $\alpha$ as the ratio of the distances $\lVert \bm{p}_{k \& l}-\bm{z}_l \rVert$ and $\lVert \bm{z}_{k}-\bm{z}_l \rVert_2$ (also refer to Fig.~\ref{fig:ratio_nonlinear}). $\alpha$ can be computed via the equation:
\begin{equation}
\alpha_{k \& l} = \frac{\lVert \bm{p}_{k \& l}-\bm{z}_l \rVert_2}{\lVert \bm{z}_{k}-\bm{z}_l \rVert_2}
= \frac{(\bm{z}_{k} - \bm{z}_{l})^T \cdot (\bm{z}_{k \& l} - \bm{z}_l)}{(\bm{z}_{k} - \bm{z}_{l})^T \cdot (\bm{z}_{k} - \bm{z}_{l})}.
\label{eq:mixing_ratio}
\end{equation}
Note that $\alpha$ lies in the range $[0, 1]$ if $\bm{z}_{k \& l}$ lies between $\bm{z}_k$ and $\bm{z}_l$, where smaller values indicate closeness to $\bm{z}_l$ and higher values indicate closeness to $\bm{z}_k$. Values smaller than zero or larger than 1 indicate that the convexity assumption is violated. Also note that the mixing ratio $\alpha$ also corresponds to the interpolation coefficient between the basic movements when we generate synthetic data.


In addition, we define the linearity Error $\delta$ as the ratio between the distance $\| \mathbf{z}_{k \& l} - \bm{p}_{k \& l} \|_2$ and the distance $\|\mathbf{z}_k - \mathbf{z}_l\|_2$, i.e.\ the distance of $\mathbf{z}_{k \& l}$ from the connecting line between $\mathbf{z}_k$ and $\mathbf{z}_l$, relative to the length of the connecting line. $\delta$ can be computed via:
\begin{equation}
\delta_{k \& l} = \frac{\| \mathbf{z}_{k \& l} - \bm{p}_{k \& l} \|_2}{\|\mathbf{z}_k - \mathbf{z}_l\|_2} = \frac{\| (\mathbf{z}_{k \& l} - \mathbf{z}_l) - \alpha_{k \& l} (\mathbf{z}_k - \mathbf{z}_l) \|_2}{\|\mathbf{z}_k - \mathbf{z}_l\|_2}.
\label{eq:linearity_error}
\end{equation}
Here, values close to zero indicate that the combined prototype $\mathbf{z}_{k \& l}$ is very close to the connecting line between the basic prototypes $\mathbf{z}_k$ and $\mathbf{z}_l$, whereas large values indicate that $\mathbf{z}_{k \& l}$ lies far away and a linear approximation may be inaccurate.
The experimental results of this analysis are shown in Sec.~\ref{sec:manifold}.


\subsection{CNN backbone}
\label{sec:backbone}
\begin{figure}
\includegraphics[width=8cm]{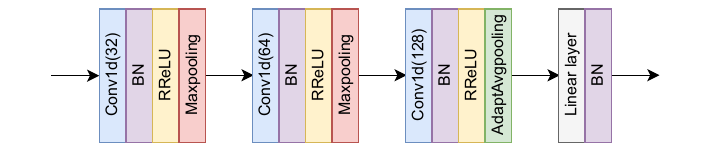}
\centering
\caption{The structure of CNN backbone.}
\label{fig:cnnbackbone}
\end{figure}

Based on prior work \cite{Liu2024,Liu2025}, we employ a 1D-CNN backbone to map input windows of multichannel sEMG signals into an embedding space. Each input time window $\bm{x}_t \in \mathbb{R}^{N \cdot C}$ where $N$ is the number of samples and $C$ is the number of channels, is mapped to a vectorial representation $\phi(\bm{x}_t) \in \mathbb{R}^n$. 
Fig.~\ref{fig:cnnbackbone} shows the CNN backbone architecture, it consists two blocks of 1D convolutions, followed by batch normalization, Random Rectified Linear Unit (RReLU) and MaxPooling, followed by a final 1D convolution, batch normalization, RReLU and average pooling, followed finally by a linear layer and batch normalization. The three 1D convolutional layers are configured with 32, 64, and 128 neurons, respectively. 
The kernel sizes are 11, 9, and 7 for the Nearlab data set, 9, 5, and 3 for our newly recorded BasCom data set, and 11, 5, 3 for amputee dataset NinaproDB3. The kernel size adjustment is chosen based on the number of samples in each time window.

\subsection{Prototype Networks}
To classify input sEMG signals into movements, we utilize prototype networks \cite{snell2017prototypical,sharma2023learning,Sato1995}. In particular, let $\phi(\bm{x}_t) \in \mathbb{R}^n$ be the embedded input signal and let $\bm{z}_1, \ldots, \bm{z}_K \in \mathbb{R}^n$ be the prototypes for the movements $1, \ldots, K$. Then, we classify the signal by computing the Euclidean distance $\lVert \phi(\bm{x}_t) - \bm{z}_k \rVert$ to all prototypes and assigning the class of the closest prototype.

To train the CNN backbone (and the prototype positions), we use the negative Euclidean distance as a logit and plug this into a cross-entropy loss:
\begin{align}
p_{t, k} &:= \frac{\exp(-\lVert \phi(\bm{x}_t) - \bm{z}_k \rVert)}{\sum_{l=1}^K \exp(-\lVert \phi(\bm{x}_t) - \bm{z}_l \rVert)} \\
\mathcal{L}_\text{crossent} &= -\sum_{t=1}^T y_{t, k} \cdot \log[p_{t, k}], \label{eq:crossent}
\end{align}
where $y_{t, k}$ is a one-hot coding of the ground-truth label for the signal time window $t$. This loss is optimized via standard deep learning techniques (Adam and backpropagation).

Importantly, we only train the prototype network on basic movements. 
In the next step, we describe our method to train prototypes for combined movements in a zero-shot, incremental learning setup.









\subsection{Compositional Prototype Interpolation (CPI):}
\begin{figure}
\includegraphics[width=8cm]{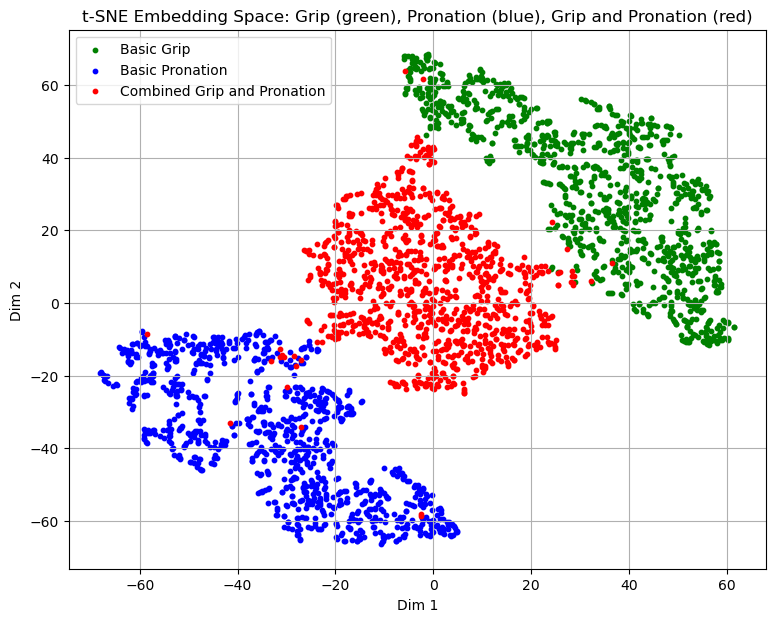}
\centering
\caption{The distribution of synthetic combined movements generated via linear interpolation in embedding space (dimensionality reduction via t-SNE).}
\label{fig:cpi}
\end{figure}

After the prototype network has been trained on seen basic movements, we wish to construct novel prototypes for pairwise combinations of movements $k$ and $l$.
To do so, we first generate synthetic samples of such combined movements via linear interpolations in the embedding space. Let $\bm{x}_k$ and $\bm{x}_l$ be two random samples from the training data for classes $k$ and $l$, respectively. Then, we construct the embedding for a combined movement $k \& l$ by uniformly sampling a mixture coefficient $\alpha \in[0, 1]$ and linearly interpolating between $\phi(\bm{x}_k)$ and $\phi(\bm{x}_l)$ via Eq.~\eqref{eq:linear_combination}.
We then initialize the prototype $\bm{z}_{k \& l}$ as the mean of all synthetic samples for for the combined movement $k \& l$. Fig.~\ref{fig:cpi} shows a t-SNE visualization of the embeddings for two basic movements (green and blue) and the synthetic data for their combination (red).





In contrast to prior work on synthetic data combinations via mixup \cite{zhang2017mixup,verma2019manifold}, our method does not require mixup during training. We keep the embedding neural network $\phi$ fixed during this process and merely construct novel prototypes in the embedding space. Hence, we keep the computational overhead of zero-shot learning extremely low and we avoid catastrophic forgetting effects for the seen basic movements. 


\subsection{Synthetic Adaptation for Prototypes (SAP)}
\label{sec:pft}

\begin{figure}
\includegraphics[width=8cm]{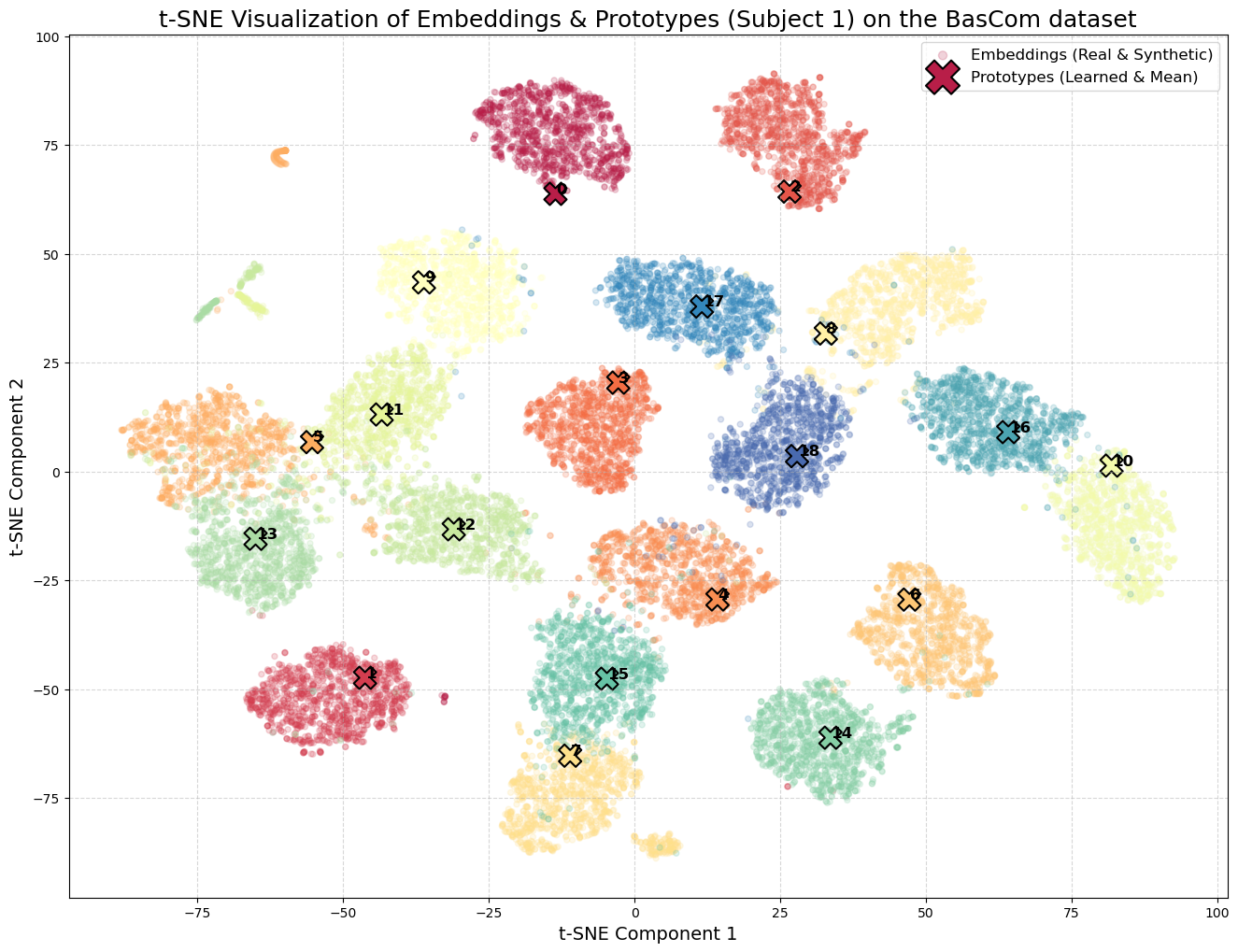}
\centering
\caption{The inconsistency between basic movement and combined movement prototypes (dimensionality reduction via t-SNE).}
\label{fig:combinedprototypes}
\end{figure}

The synthetic generated embeddings can be further utilized to adapt the prototypes for classification, without any need for newly recorded sEMG data (meaning: this remains a zero-shot method).
In CPI, the prototypes of combined movements are computed as the mean of synthetic data, which may be suboptimal for classification and inconsistent with the learned prototypes for basic movements. Fig.~\ref{fig:combinedprototypes} illustrates this inconsistency: The learned prototypes for basic movements are sometimes at the margins of their respective class to optimize decision boundaries, whereas the prototypes of combined movements remain (by construction) in the mean, leading to misclassifications.

We investigate two zero shot adaptation methods to optimize the locations of prototypes with synthetic data.


\paragraph{SAP-full:} In the SAP-full method, we simply concatenate the training data for basic movements with the synthetic training data for combined movements and then adapt the location of all prototypes (for basic and combined movements) to minimize the crossentropy loss~\eqref{eq:crossent}. Importantly, we do not train the CNN backbone $\phi$ at this point, such that learning occurs purely in the embedding space and is thus an efficient, convex optimization problem \cite{Paassen2018Neurocomputing}.




\paragraph{SAP-Specialized:} SAP-full optimizes the overall crossentropy loss across all classes. However, this may overemphasize the accuracy on basic movements, which are easier to classify. For amputees in particular, the embedding space tends to be more distorted and non-linear (refer to Section~\ref{sec:manifold} for experimental results), such that additional algorithmic changes may be needed. We propose a combination of three loss functions:



\begin{equation}
\mathcal{L}_{\text{total}} 
= \mathcal{L}_{\text{crossent}} 
+ \lambda_m \cdot \mathcal{L}_{\text{margin}} 
+ \lambda_s \cdot \mathcal{L}_{\text{stability}},
\end{equation}
where the $\lambda_m$ and $\lambda_s$ are hyper-parameters to emphasize different parts of the loss.
The crossentropy loss $\mathcal{L}_{\text{crossent}}$ from Eq.~\eqref{eq:crossent} is computed only for the combined movements, forcing the optimization to put more emphasis on them. The term $\mathcal{L}_{\text{margin}}$ is intended to push the prototypes of basic and combined movements apart and is defined as:
\begin{equation}
    \mathcal{L}_{\text{margin}}
    = \frac{1}{N_B \cdot N_C}
    \sum_{b=1}^{N_B} \sum_{c=1}^{N_C}
    \exp\!\left(-\frac{\lVert \bm{z}^{(b)}_{\text{basic}} - \bm{z}^{(c)}_{\text{comb}} \rVert}{\tau}\right), \label{eq:margin}
\end{equation}
where $N_B$ and $N_C$ are the number of basic movement prototypes and combined movement prototypes, respectively, and $\tau$ is a learnable parameter that adaptively adjusts the separation scale, finding the optimal inter-class interval to separate the prototypes.

The third term, $\mathcal{L}_{\text{stability}}$, is a stability loss to prevent the basic movement prototypes from moving too far from their original positions, thus preventing catastrophic forgetting.
\begin{equation}
\mathcal{L}_{\text{stability}} 
:= \frac{1}{N_B} 
\sum_{b=1}^{N_B} 
\left\| \bm{z}_{basic}^{\text{(b, after)}} 
- 
\bm{z}_{basic}^{\text{(b, before)}} 
\right\|_2^2. \label{eq:stability}
\end{equation}  


\subsection{Extended threshold-based zero-shot classification}
\label{sec:extended}

As baseline for our experiments, we consider the zero-shot classification methods Syn0net proposed by \cite{soroushmojdehi2022transfer}. The Syn0net inference scheme for zero-shot classification assumes a linear layer for movement classification, then takes the two most activated movements $k$ and $l$ and predicts their combination $k \& l$. However, this method \emph{always} predicts combined movements and can, hence, not be used in an incremental learning setup where basic movements should also be predicted. We extend their scheme in two ways:

\begin{figure}
\includegraphics[width=8cm]{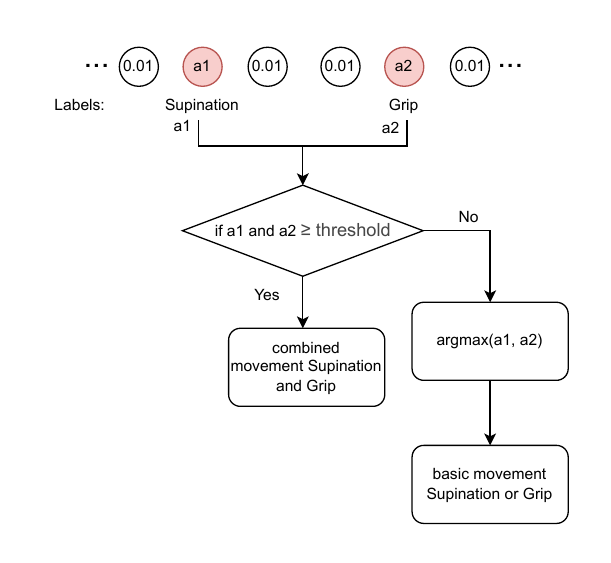}
\centering
\caption{The new inference framework for zero-shot incremental learning} 
\label{fig:threshold}
\end{figure} 

\paragraph{Extended inference schema (e-Syn0net):} Similar to \cite{soroushmojdehi2022transfer}, we investigate the maximum activation among all hand movements $k$ and among all wrist movements $l$. If either is below a threshold, we predict a basic movement (namely the most activated movement). However, if both are above a threshold, we predict their combination $k \& l$ (see Fig.~\ref{fig:threshold}).





\paragraph{Signal mixup:} Simply selecting a threshold between basic and combined movements may be insufficient for good classification accuracy on both basic and combined movements because the model is only trained on basic movements. Accordingly, we also investigate a signal mixup scheme \cite{zhang2017mixup}, where synthetic training data for the combined movement $k \& l$ is generated via linear interpolation in the input space. In particular, we randomly sample training data points $\bm{x}_k$ and $\bm{x}_l$ as well as an interpolation factor $\alpha \in [0, 1]$ and then construct a combined training data point $\bm{x}_{k \& l}$ as well as its label $y_{k \& l}$ via:
\begin{align}
\bm{x}_{k \& l} &\approx \alpha \cdot \bm{x}_{k} + (1-\alpha) \cdot \bm{x}_l \\
y_{k \& l} &\approx \alpha \cdot y_k + (1-\alpha) \cdot y_l,
\end{align}
yielding a data set with soft labels. Then, the neural net (the CNN as well as the final classification layer) is trained on this combined training data set via the crossentropy loss.



We acquire a novel data set and perform online evaluation experiments via a user study, which we describe in the following.


\subsection{Participants information}
We recruited 11 participants, 5 female and 6 male, via a convenience sample of students and researchers at Bielefeld University. The metadata of all participants is provided in Tab.~\ref{tab:participants}. All participants provided informed consent after receiving a participant information sheet. The study was approved by the Bielefeld University ethics review board under application number 2025-304 on August 05, 2025.


\begin{center}
\begin{table}[htbp]
\centering
\caption{Demographic information of participants.}
\begin{tabular}{c c c c}
\hline
\textbf{ID} & \textbf{Age (years)} & \textbf{Gender} & \textbf{Handedness} \\ 
\hline
P1 & 30-40 & Male & Right \\
P2 & 20-30 & Female & Right \\
P3 & 20-30 & Male & Right \\
P4 & 20-30 & Female & Right \\
P5 & 20-30 & Female & Right \\
P6 & 20-30 & Female & Right \\
P7 & 20-30 & Male & Right \\
P8 & 20-30 & Male & Right \\
P9 & 20-30 & Female & Right \\
P10 & 20-30 & Male & Right \\
P11 & Missing & Female & Right \\
\hline
\end{tabular}
\label{tab:participants}
\end{table}
\end{center}

\subsection{Hardware}
The sEMG signal is collected by using MindRove armband (MindRove, https://www.mindrove.com), which collects sEMG signals around the forearm with 8 channels.
The sampling frequency is 500Hz, resolution is 24 bit and data transmission is realized via WIFI and the Mindrove SDK.
For our dataset, the data acquisition and real-time inference is done on a Laptop with AMD Ryzen 5 5500U with Radeon Graphics, 8 GB RAM. To enable fast training times during the study, model training is performed via the Kaggle Cloud platform with a Intel(R) Xeon(R) CPU @ 2.00GHz and Nvidia Tesla P100-PCIE-16GB. 

\subsection{User Interface}

\begin{figure}
\includegraphics[width=7cm]{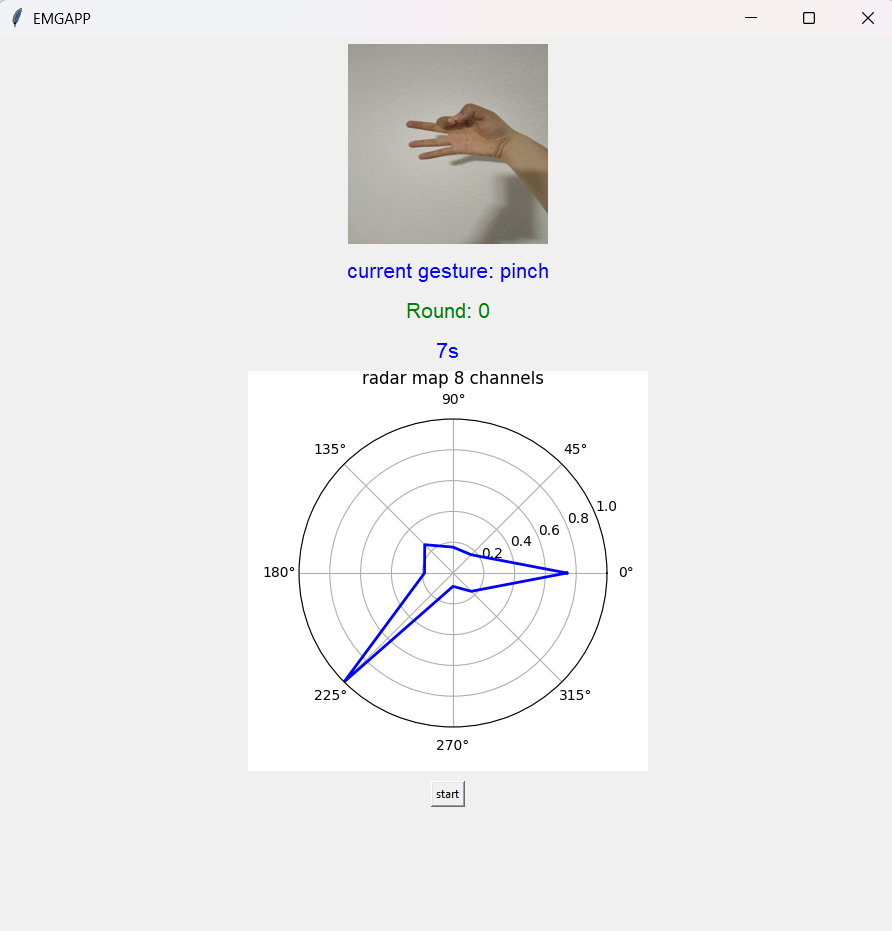}
\centering
\caption{The radar chart of 8 sEMG channels}
\label{fig:radar}
\end{figure}

The user interface for the study is programmed in Python and Tkinter. During training data acquisition, the movement that shall be performed and the current signal is shown to participants (Fig.~\ref{fig:radar}). The signal is visualized as a radar chart, displaying the current sEMG amplitude in each of the 8 channels.

During online evaluation, the currently desired movement as well as the prediction of the model is shown (Fig.~\ref{fig:inference}).

\subsection{Study protocol}
\label{sec:protocol}

Participants are first asked to collect data for both basic and combined movements.
The basic movements include 4 wrist movements (extension, flexion, supination, and pronation) as well as 6 hand movements (pinch, lateral pinch, tripod, hook, hand open and grip), and rest. The combined movements are 8 pairwise combinations of wrist and hand movements, namely extension \& open, extension \& hook, extension \& pinch, flexion \& pinch, pronation \& hook, supination \& grip, supination \& open, and supination \& hook (Fig.~\ref{fig:movements}). The combined movements are chosen to align with everyday activities, such as pulling out a drawer or opening a door. Overall, we cover 11 basic and 8 combined movements, so 19 movements overall.

\begin{figure}
\includegraphics[width=4cm]{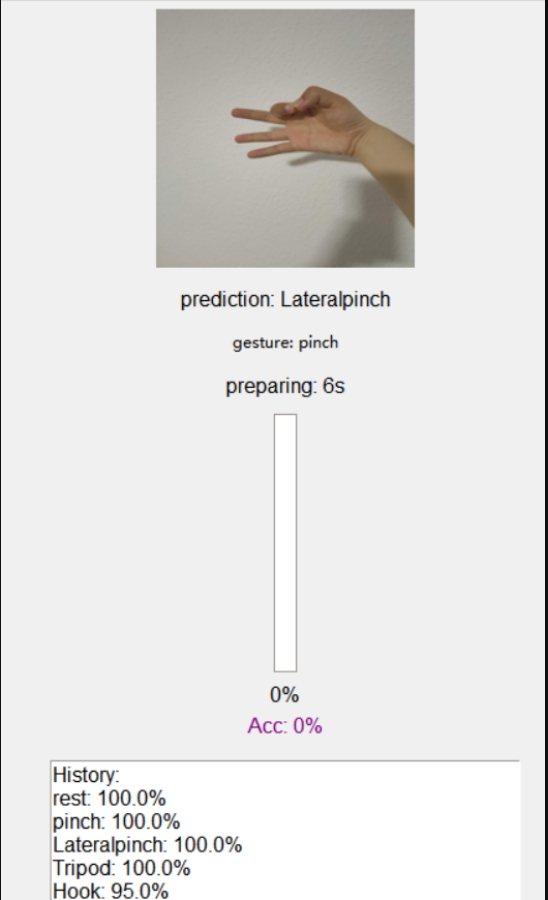}
\centering
\caption{The GUI of inference process}
\label{fig:inference}
\end{figure}

During online evaluation, the currently desired movement as well as the prediction of the model is shown (Fig.~\ref{fig:inference}). 

\begin{figure}
\includegraphics[width=8cm]{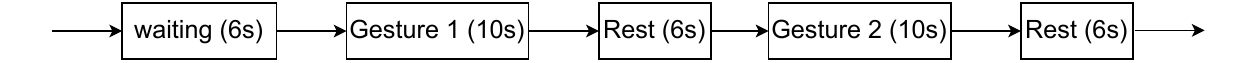}
\centering
\caption{An illustration of the data collection process}
\label{fig:datacollectionprocess}
\end{figure}

The protocol begins with a waiting phase for 6 seconds in which no data is recorded and in which participants can prepare for the first movement. Participants are asked to perform the movement for 10 seconds, during which data is recorded, then rest for 6 seconds, which is recorded as well (Fig.~\ref{fig:datacollectionprocess}). Movement and rest phases are interleaved until all movements are recorded. This is repeated 3 times for all basic and combined movements, then 7 more times for only the basic movements.
Based on the recorded data, three models are trained, namely e-Syn0net, e-Syn0net + signal mixup and our proposed SAP-full scheme.

Then, the online evaluation is performed. For each movement (basic and combined), participants have 6 seconds to practice the movement and observe the prediction of the model, after which the predictions are recorded for 4 seconds.

\section{Experiments}
We compare the performance of baseline and proposed models on three data sets offline (Nearlab, BasCom, our new data set, and NinaPro DB3), as well as in an online user study. 


\subsection{Datasets}
For our offline evaluation, we use three datasets: Nearlab, our newly recorded BasCom data set, and NinaPro DB3.

\paragraph{Nearlab} The Nearlab dataset includes sEMG data from 11 participants performing 8 basic movements (flexion, extension, supination, pronation, hand open, pinch, lateral pinch, and grip) and 6 combined movements (pronation or supination combined with pinch, lateral pinch, and grip). Each basic movement is repeated 5 times in 3 different positions of the hand, while 6 combined movements are repeated 5 times in one position. 10 differential sEMG channels are recorded by electrode pairs placed around the forearm. The sampling frequency is  2,048 Hz. The dataset is pre-processed with a 10-500 Hz bandpass filter and a 50 Hz notch filter, segmented into 512-sample windows with strides of 128 samples. 
In our experiment, we split the basic movement data according to different repetitions in the dataset. The first and second basic movement rounds are used for training, while the third round is for testing. Furthermore, all combined movements test the dataset. 

\begin{figure}
\includegraphics[width=8cm]{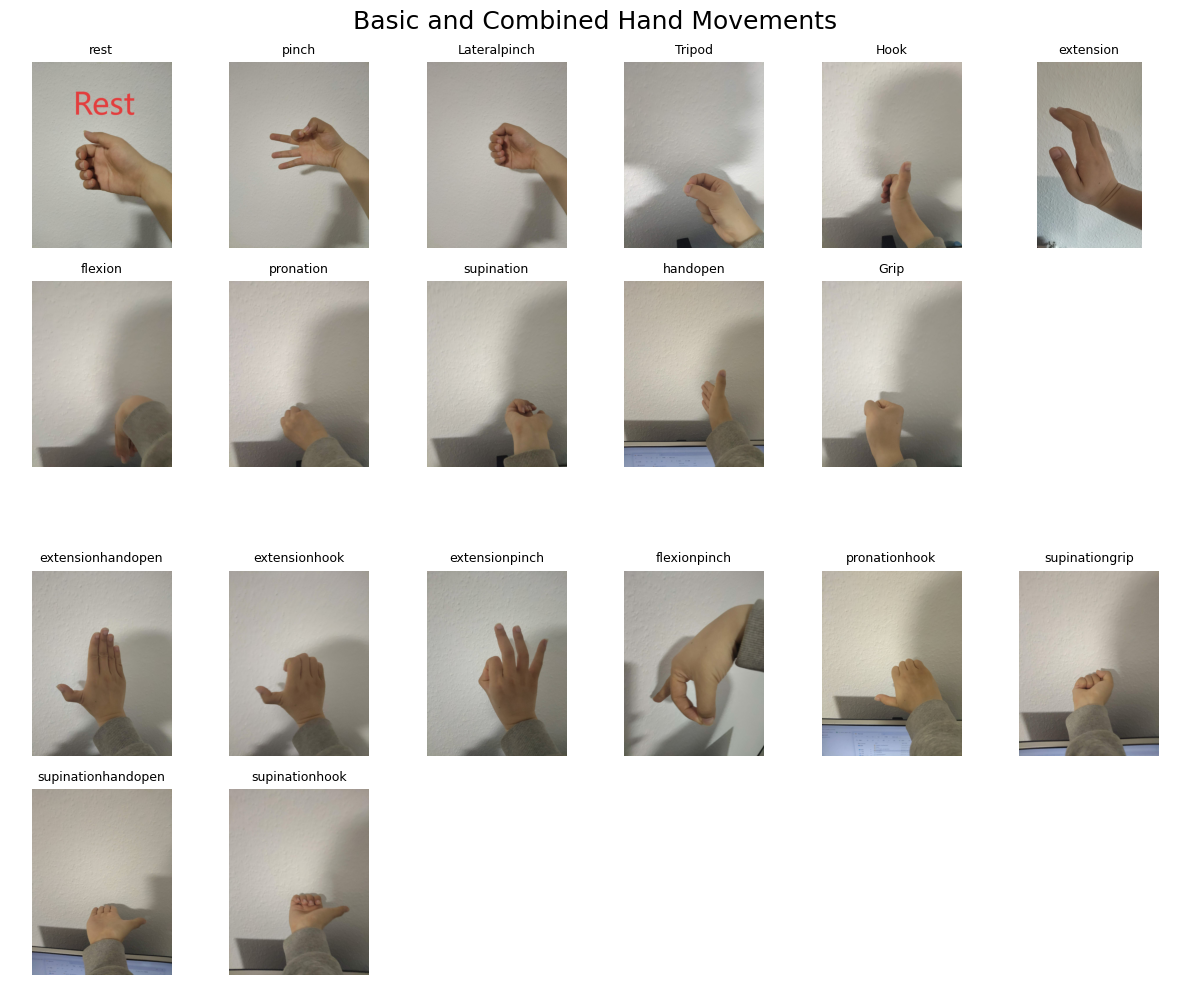}
\centering
\caption{The basic and combined movements included in the study}
\label{fig:movements}
\end{figure}

\paragraph{BasCom} Our BasCom dataset contains 19 different movements, consisting of 11 basic movements and 8 combined movements (Fig.~\ref{fig:movements}). 
The dataset consists of 10 round repetitions; the first 3 rounds have all 19 movements, and the last 7 rounds contain only 11 basic movements. The dataset is collected by Mindrove, an armband placed on the forearm. Its sampling frequency is 500 Hz with 8 channels. The signals are filtered with a 20-250 Hz bandpass filter and a 50 Hz notch filter and then segmented into 128-sample windows with a sliding step size of 25 samples.

\paragraph{NinaPro DB3} NinaPro DB3 dataset is a benchmark database for investigating sEMG recognition in amputees \cite{atzori2014electromyography}. The NinaPro DB3 dataset contains 11 amputees with different percent remaining forearm and experience of prosthesis.
The NinaPro DB3 dataset includes 3 exercises: Exercise B (9 basic wrist motions and 8 isometric and isotonis hand configuration), Exercise C (23 grasping and functional motions) and Exercise D (9 force patterns). 
For our task, we selected 15 motions to evaluate our models.
The seen motion set consists of 14 seen classes, separated into training and test datasets according to repetitions:
\begin{itemize}
\item Rest
\item Basic motions: 7 basic hand from the Exercise B, such as Fist, Wrist Flextion and Wrist Extension.
\item Force Patterns: 6 force patterns selected from the Exercise D.
\end{itemize}

For unseen set, we selected the "Wrist Extension with Closed Hand" motion from Exercise B. This motion can be viewed as the combination of two basic motions: "Fist" and "Wrist Extension". This combined motion is completely excluded from the training. Across data set, the "Rest" movement is subsampled to be balanced with the other movement classes.

\subsection{Models}

For all models, we use the same 1D-CNN backbone as described in Sec.~\ref{sec:backbone} but different classification layers, training, and inference schemes. To our best knowledge, only a few relevant sEMG studies have addressed to recognize unseen combined movements using only seen basic movements.

\paragraph{Syn0net} is the zero-shot classification scheme of \cite{soroushmojdehi2022transfer}. The model uses a linear classification layer trained via crossentropy loss on seen basic movements. For inference in the incremental setting (where seen basic and unseen combined movements occur), we use the extended inference scheme (e-Syn0net) described in Sec.~\ref{sec:extended}.

\paragraph{KL inference} is the method proposed by \cite{yazawa2025recognition} but with the CNN backbone instead of an MLP. The model uses a classification layer for basic movements but the inference is performed via the KL divergence between the prediction and target label vectors for each single class and class combination. The training is performed by combining signals for basic movements with synthetic combined signals via linear interpolation in the input space (refer to Sec.~\ref{sec:extended}).

\paragraph{e-Syn0net+Mixup} is e-Syn0net (see above) combined with a training on synthetic combined movements as described in  Sec.~\ref{sec:extended}.

\paragraph{SAP-full} is our proposed prototype network architecture where prototypes for combined movements are initialized as the mean of synthetic data in the embedding space (CPI) and then all prototypes are adapt (SAP) to reduce the crossentropy loss while keeping the CNN backbone fixed, as described in Sec.~\ref{sec:pft}.

\paragraph{SAP-specialized} refers to different variants of prototype adaptation as described in Sec.~\ref{sec:pft}.

To train the models, we use 100 epochs of AdamW with a learning rate of $0.001$, weight decay of $0.001$, and minibatch size of 32. Training on the Kaggle platform ran within 5 minutes for all models. Offline experiments were repeated 5 times with different random seeds to enhance robustness of results.

\subsection{Geometric Analysis}
\label{sec:manifold}

\begin{figure}
\includegraphics[width=8cm]{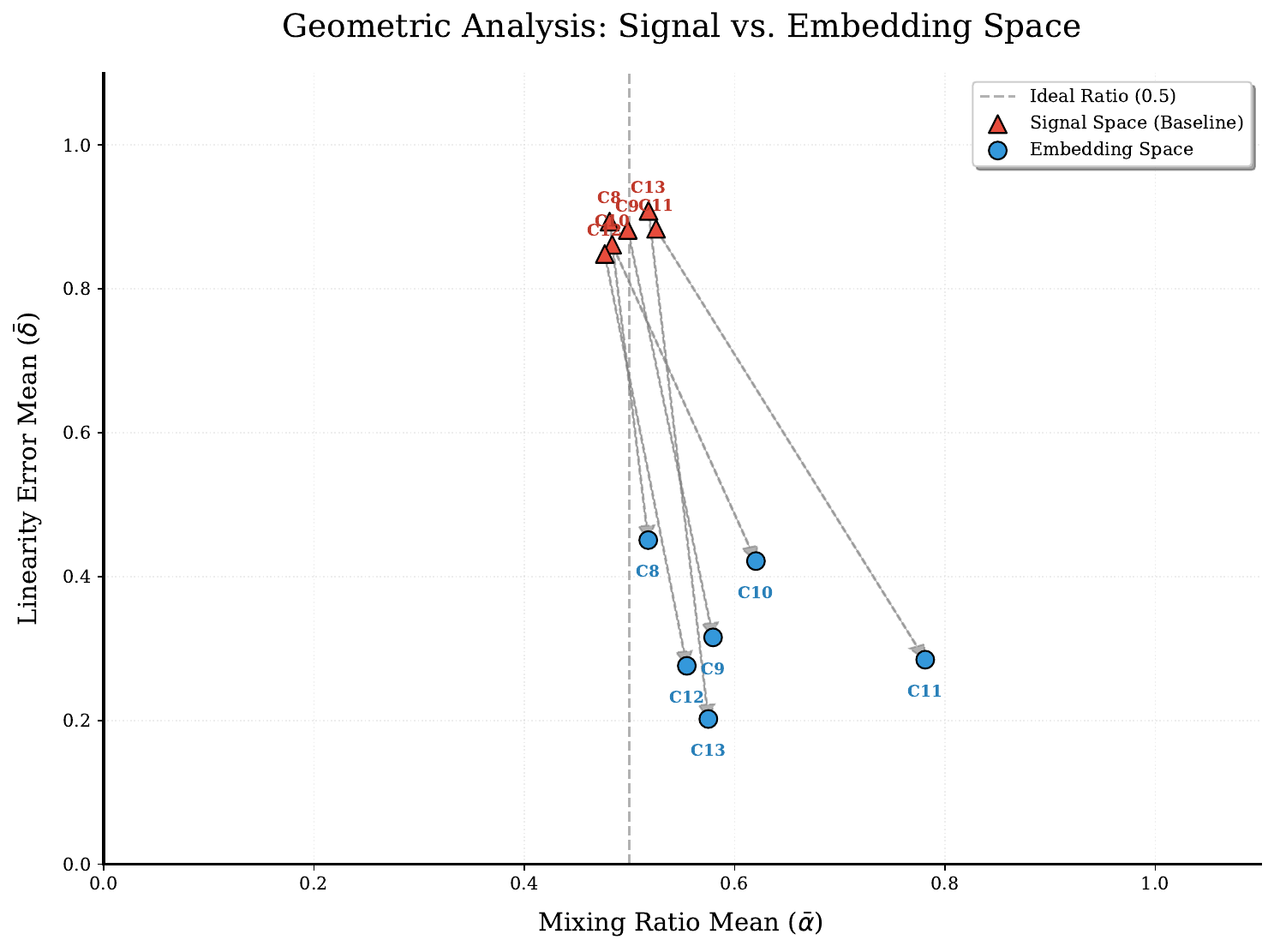} 
\centering
\caption{The geometric analysis on Nearlab.} 
\label{fig:linearization}
\end{figure}

Before presenting classification results, we study the geometric relation of unseen combined and seen basic movements using the aforementioned mixing ration $\alpha$ from Eq.~\eqref{eq:mixing_ratio} and the linearity error $\delta$ from Eq.~\eqref{eq:linearity_error}. Fig.~\ref{fig:linearization} shows $\alpha$ on the x axis and $\delta$ on the y axis for the Nearlab data set. Each marker represents one combined prototype. Triangles indicate prototypes in the signal space, circles in the embedding space. We observe that mixing ratios $\alpha$ always lie well between 0 and 1, as desired. However, the linearity error $\delta$ is notably higher in signal space with values above $0.8$, whereas values are smaller $0.5$ in the embedding space, indicating that the prototype network is able to flatten the data manifold to some degree and suggesting that a liner interpolation of combined movements between basic movements is possible.


Values are notably worse for amputees (NinaPro DB3 data set), as shown in Table~\ref{tab:geometry_comparison}. Here, the median linearity error remains relatively high at $0.76$ even in embedding space. Further, mixing ratios are closer to the extremes with $0.39$ in signal and $0.78$ in the embedding space, indicating that one type of movement (either hand or arm) dominates combines movements.
\begin{table}[h]
\centering
\caption{Mean linearity error ($\delta$) and mixing ratio ($\alpha$) across data sets for both signal and embedding space. 
}
\label{tab:geometry_comparison}
\begin{scriptsize}
\begin{tabular}{l l c c c c} 

\multirow{2}{*}{\textbf{Dataset}} & 
\multirow{2}{*}{\textbf{Subject Type}} & 
\multicolumn{2}{c}{\textbf{Signal Space}} & 
\multicolumn{2}{c}{\textbf{Embedding Space}} \\

 & & $\delta_{raw}$ & $\alpha_{raw}$ & $\delta_{emb}$ & $\alpha_{emb}$ \\ 
\cmidrule(lr){1-1} \cmidrule(lr){2-2} \cmidrule(lr){3-4} \cmidrule(lr){5-6}
Nearlab         & Able-bodied               & 0.88   & 0.50     & 0.32   & 0.60 \\
BasCom          & Able-bodied               & 0.81   & 0.49     & 0.29   & 0.65 \\ 
\cmidrule(lr){1-1} \cmidrule(lr){2-2} \cmidrule(lr){3-4} \cmidrule(lr){5-6}
NinaPro DB3$^*$ & Amputee               & 0.83   & 0.39     & 0.76   & 0.78 \\ 
\multicolumn{6}{l}{\footnotesize{$^*$Reported as median values to counteract the impact of outliers.}}
\end{tabular}
\end{scriptsize}
\end{table}

\subsection{Zero-shot classification}

To evaluate zero-shot classification performance, we first train models on seen basic movements and evaluate them on unseen combined movements. Table~\ref{tab:results_combined} shows the results, indicating that SAP-full outperforms all baselines on the Nearlab data set and performs comparably to the best model, KL inference, on the BasCom data set. However, this evaluation only refers to unseen combined movements, whereas any practical system would still need to perform well on basic movements, as well, meaning an incremental learning setup.


\begin{table}
\caption{Mean accuracy ($\pm$ std.) across participants of all methods on the Nearlab and BasCom datasets (only contains unseen movements for evaluation). }
\label{tab:results_combined}
\centering
\begin{scriptsize}
\begin{tabular}{lcc}
\textbf{Method} & \textbf{Nearlab} & \textbf{BasCom} \\
\cmidrule(lr){1-1} \cmidrule(lr){2-3}
Syn0net & $0.67\pm 0.13$ & $0.49\pm 0.07$ \\
KL inference & $0.73\pm 0.09$ & $\bm{0.66\pm 0.08}$ \\
Syn0net+Signal mixup & $0.72 \pm 0.09$ & $0.59 \pm 0.09$  \\
SAP-full & $\bm{0.75\pm 0.11}$ & $0.65 \pm 0.07$\\
\end{tabular}
\end{scriptsize}
\end{table}

\begin{table}
\caption{Mean accuracy ($\pm$ std.) across participants of all methods on the Nearlab dataset}
\label{tab:results_nearlab}
\centering
\begin{scriptsize}
\begin{tabular}{lccc}
\textbf{Method} & \textbf{Seen} & \textbf{Unseen} & \textbf{All} \\
\cmidrule(lr){1-1} \cmidrule(lr){2-4}
e-Syn0net & $\bm{0.78\pm 0.08}$ & $0.12\pm 0.05$ & $0.51\pm 0.06$ \\
KL inference &$0.74\pm 0.07$ & $0.27\pm 0.13$ & $0.55\pm 0.09$ \\
e-Syn0net+Signal mixup & $0.70 \pm 0.08$ & $0.36 \pm 0.16$ & $0.56\pm 0.10$ \\
SAP-full & $0.63\pm 0.07$ & $\bm{0.51 \pm 0.12}$ & $\bm{0.58\pm 0.08}$ \\
\end{tabular}
\end{scriptsize}
\end{table}

Tab.~\ref{tab:results_nearlab} shows the results when seen basic movements are included and where the extended inference scheme for Syn0net is needed. In this case, Syn0net performs best for the seen basic movements but accuracy breaks down dramatically for unseen combined movements, where SAP-full performs best. SAP-full retains much more accuracy in the zero-shot incremental setting compared to the baselines. We also note that the performance difference between SAP-full and the next-best model (e-Syn0net+Signal mixup) on unseen combined movements is significant ($p < 0.005$ in a Wilcoxon signed-rank test) in the zero-shot incremental setting. 

We observe the same qualitative pattern for the BasCom data set. Tab.~\ref{tab:results_bascom} shows the results, where SAP-full, once more, performs best for unseen combined and all movements in the zero-shot incremental setting. The difference to the second-best model (KL inference) is significant on both combined movement and all movement ($p < 0.0005$ in a Wilcoxon signed-rank test).

\begin{table}
\caption{Mean accuracy ($\pm$ std.) across participants of all methods on BasCom dataset}
\label{tab:results_bascom}
\centering
\begin{scriptsize}
\begin{tabular}{lccc}
\textbf{Method} & \textbf{Seen} & \textbf{Unseen} & \textbf{All} \\
\cmidrule(lr){1-1} \cmidrule(lr){2-4}
e-Syn0net & $\bm{0.86\pm 0.05}$ & $0.09\pm 0.04$ & $0.54\pm 0.04$ \\
KL inference & $\bm{0.86\pm 0.06}$ & $0.17\pm 0.05$ & $0.57\pm 0.05$ \\
e-Syn0net+Signal mixup & $0.82 \pm 0.06$ & $0.21 \pm 0.05$ & $0.56\pm 0.05$ \\
SAP-full & $0.81\pm 0.07$ & $\bm{0.40 \pm 0.09}$ & $\bm{0.64\pm 0.06}$ \\
\end{tabular}
\end{scriptsize}
\end{table}




While SAP-full does perform best across methods, we still observe low accuracy in the zero-shot incremental setting for unseen combined movements. Hence, we explore further adaptation variants to resolve this problem.

\subsubsection{Synthetic Adaptation experiments}

To evaluate the impact of different synthetic adaptation methods, we compare the initialized prototypes without adaptation (CPI) with SAP-full and SAP-specialized using only the margin loss from Eq.~\eqref{eq:margin} and SAP-specialized using both the margin and stability loss from Eq.~\eqref{eq:stability}.

Tab.~\ref{tab:results_ft_nearlab} shows the results on the Nearlab data set. We observe that no adaptation yields the best results for seen basic movements but SAP-full yields the best results for all movements. However, if a more balanced accuracy between seen basic and unseen combined movements is desired, SAP-specialized with margin and stability loss may be the method of choice. The difference between SAP-Specialized and CPI on combined movements is significant ($p < 0.0005$).

\begin{table}
\caption{Mean accuracy ($\pm$ std.) across participants of adaptation methods on the Nearlab dataset}
\label{tab:results_ft_nearlab}
\centering
\begin{scriptsize}
\begin{tabular}{lccc}
\textbf{Method} & \textbf{Seen} & \textbf{Unseen} & \textbf{All} \\
\cmidrule(lr){1-1} \cmidrule(lr){2-4}
CPI & $\bm{0.65\pm 0.08}$ & $0.43\pm 0.10$ & $0.56\pm 0.08$ \\
\cmidrule(lr){1-1} \cmidrule(lr){2-4}
SAP-full & $0.63\pm 0.07$ & $0.51\pm 0.11$ & $\bm{0.58\pm 0.08}$ \\
SAP-Spec.\ w/o $\mathcal{L}_{\text{stability}}$ & $0.47 \pm 0.09$ & $\bm{0.58 \pm 0.10}$ & $0.52\pm 0.09$ \\
SAP-Specialized & $0.55 \pm 0.09$ & $0.56 \pm 0.11$ & $0.55\pm 0.09$ \\
\end{tabular}
\end{scriptsize}
\end{table}



Tab.~\ref{tab:results_ft_bascom} shows a similar pattern for the BasCom data set. Here, SAP-full achieves the best results both for seen basic and all movements, whereas SAP-specialized without stability loss achieves the highest accuracy for unseen combined movements only. SAP-specialized with both margin and stability loss achieves significantly better results on combined movements compared to CPI ($p < 0.0005$) and may offer a compromise in maintaining reasonable accuracy on basic movements as well as combined movements.

\begin{table}
\caption{Mean accuracy ($\pm$ std.) across participants of adaptation methods on BasCom dataset}
\label{tab:results_ft_bascom}
\centering
\begin{scriptsize}
\begin{tabular}{lccc}
\textbf{Method} & \textbf{Seen} & \textbf{Unseen} & \textbf{All} \\
\cmidrule(lr){1-1} \cmidrule(lr){2-4}
CPI & $0.74\pm 0.11$ & $0.46\pm 0.08$ & $0.62\pm 0.06$ \\
\cmidrule(lr){1-1} \cmidrule(lr){2-4}
SAP-full & $\bm{0.81\pm 0.07}$ & $0.40\pm 0.09$ & $\bm{0.64\pm 0.06}$ \\
SAP-Spec.\ w/o $\mathcal{L}_{\text{stability}}$ & $0.58 \pm 0.11$ & $\bm{0.57 \pm 0.08}$ & $0.58\pm 0.08$ \\
SAP-Specialized & $0.66 \pm 0.12$ & $0.53 \pm 0.07$ & $0.61\pm 0.08$ \\
\end{tabular}
\end{scriptsize}
\end{table}

\subsubsection{Amputee Evaluation}
Both NearLab and BasCom are limited to able-bodied participants. As shown in Sec.~\ref{sec:manifold}, the embedding space for amputees tends to be more non-linear, suggesting that zero-shot movement classification may be more difficult for amputees. Therefore, we also evaluate on the NinaProDB3 dataset. We select 14 basic movements as training data and consider the combined movement "Wrist Extension with Closed Hand" (as combination of "Fist" and "Wrist Extension") to evaluate zero-shot classification performance.

\begin{table}
\caption{Median accuracy (IQR) across amputees}
\label{tab:results_amputee}
\centering
\begin{scriptsize}
\begin{tabular}{lccc}
\textbf{Method} & \textbf{Seen} & \textbf{Unseen} & \textbf{All } \\
\cmidrule(lr){1-1} \cmidrule(lr){2-4}
KL Inference & $0.48(0.18)$ & $0.04(0.05)$ & $0.46(0.18)$ \\
e-Syn0net + Mixup & $0.48(0.20)$ & $0.01(0.06)$ & $0.47(0.20)$ \\
\cmidrule(lr){1-1} \cmidrule(lr){2-4}
SAP-full & $0.51 (0.22)$ & $0.17 (0.20)$ & $0.50 (0.20)$\\
SAP-Specialized & $0.44 (0.22)$ & $\bm{0.43 (0.30)}$ & $0.45 (0.21)$
\end{tabular}
\end{scriptsize}
\end{table}

Tab.~\ref{tab:results_amputee} presents the evaluation results.
The baseline methods relying on signal mixup, such as KL inference and e-Syn0net, fail to recognize unseen combined movements.
Even SAP-full achieves only 14\% median accuracy on combined movements. 
By contrast, the SAP-specialized achieves 43\% median accuracy. 
This confirms the effectiveness of our proposed SAP-specialized method, which is designed for recognizing unseen combined motion in challenging environments.

\subsubsection{Online Evaluation}

To gauge the performance of models in a more realistic setup where real-world disturbances may occur, we also performed online movement classification with e-Syn0net, e-Syn0net + Signal Mixup and SAP-full following the protocol described in Sec.~\ref{sec:protocol}. 

Tab.~\ref{tab:results_online} shows the results of the experiments. We observe essentially the same pattern as in the offline incremental zero-shot classification setup: e-Syn0net achieves better results on seen basic movements, but SAP-full performs much better on unseen combined and all movements.

\begin{table}
\caption{Mean accuracy ($\pm$ std.) across participants of Real-time inference with Mindrove armband}
\label{tab:results_online}
\centering
\begin{scriptsize}
\begin{tabular}{lccc}
\textbf{Method} & \textbf{Seen} & \textbf{Unseen} & \textbf{All} \\
\cmidrule(lr){1-1} \cmidrule(lr){2-4}
e-Syn0net & $\bm{0.88\pm 0.07}$ & $0.08\pm 0.06$ & $0.54\pm 0.05$ \\
\cmidrule(lr){1-1} \cmidrule(lr){2-4}
e-Syn0net + Mixup & $0.80\pm 0.09$ & $0.19\pm 0.09$ & $0.55\pm 0.07$ \\
SAP-full & $0.76 \pm 0.08$ & $\bm{0.47 \pm 0.11}$ & $\bm{0.63\pm 0.08}$ 
\end{tabular}
\end{scriptsize}
\end{table}

\section{Discussion and Conclusion}

In this work, we proposed a new scheme to perform zero-shot classification for unseen combined movements from sEMG data. We began with a geometric analysis method to verify that the embedding space derived from a prototype network makes the relationship between unseen combined and seen basic movements more linear compared to signal space.
Thereby, we have established a foundation for our approach of approximating unseen combined movements via linear interpolation in the embedding space.
To arrive at a zero-shot classifier for unseen combined movements, we trained a prototype network on basic movements only and then interpolated the prototypes for combined movements from the basic movements in embedding space (which we call compositional prototype interpolation, CPI). Further, we adapt these prototypes in the embedding space based on synthetic, interpolated training data points for combined movements (synthetic adaptation for prototypes). We showed that the resulting network outperforms prior zero-shot classification schemes for combined movements by a large margin across data sets: On able-bodied subjects in both the pre-existing NearLab and our novel BasCom data set, on amputees in the NinaPro DB3 data set, and in an online user study. As such, our results suggest that prototype networks have a strong potential as zero-shot classifiers for combined movements.

\subsection{Limitations}
Although we achieved notable successes with prototype networks, some limitations persist. First, we still observe a tradeoff between basic and combined movement accuracy, where the latter can only be enhanced at expense of the former. Second, due to the CNN backbone, the prototypes themselves lack interpretability: we can not directly show prototypes in the input space. Further explainability methods are required. Finally, our online evaluation only includes able-bodied participants, such that pre-clinical and clinical trials for zero-shot sEMG movement classification with amputees would still be required.

\bibliographystyle{plainnat}
\bibliography{literature}
\end{document}